\documentclass[10pt,twocolumn,letterpaper]{article}

\usepackage{iccv}
\usepackage{times}
\usepackage{epsfig}
\usepackage{graphicx}
\usepackage{amsmath}
\usepackage{amssymb}

\usepackage{algorithm}
\usepackage{algorithmic}
\usepackage{booktabs}       
\usepackage{amsfonts}       
\usepackage{nicefrac}       
\usepackage{microtype}      
\usepackage{multirow}
\usepackage{easyReview}
\usepackage{latexsym}
\usepackage{amsmath}

\usepackage{subcaption}

\usepackage{xcolor}

\usepackage[breaklinks=true,bookmarks=false]{hyperref}

\iccvfinalcopy 

\def\iccvPaperID{****} 

\begin{document}

\title{Multi-Modal Hybrid Learning and Sequential Training for RGB-T Saliency Detection}

\author{Guangyu Ren\\
{\tt\small r.guangyu@ucl.ac.uk}
\and
Jitesh Joshi\\
{\tt\small jitesh.joshi.20@ucl.ac.uk}
\and
Youngjun Cho\\
{\tt\small youngjun.cho@ucl.ac.uk}
}

\maketitle

\begin{abstract}
RGB-T saliency detection has emerged as an important computer vision task, identifying conspicuous objects in challenging scenes such as dark environments. However, existing methods neglect the characteristics of cross-modal features and rely solely on network structures to fuse RGB and thermal features. To address this, we first propose a Multi-Modal Hybrid loss (MMHL) that comprises supervised and self-supervised loss functions. The supervised loss component of MMHL distinctly utilizes semantic features from different modalities, while the self-supervised loss component reduces the distance between RGB and thermal features. We further consider both spatial and channel information during feature fusion and propose the Hybrid Fusion Module to effectively fuse RGB and thermal features. Lastly, instead of jointly training the network with cross-modal features, we implement a sequential training strategy which performs training only on RGB images in the first stage and then learns cross-modal features in the second stage. This training strategy improves saliency detection performance without computational overhead. Results from performance evaluation and ablation studies demonstrate the superior performance achieved by the proposed method compared with the existing state-of-the-art methods. 
\end{abstract}

\section{Introduction}
\label{sec:intro}
Salient object detection simulates the human visual attention mechanism that identifies or segments prominent objects in a given scene. Saliency detection methods relying solely on RGB images often fail to segment objects in some challenging scenes where objects are affected by poor lighting conditions. In such cases, multi-modal imaging can be leveraged with each modality complementing the other. One such complementary modality is thermal infrared imaging which does not depend on ambient lighting conditions \cite{joshi2022self}. RGB-T saliency detection can build upon the advancements in RGB as well as RGB-T semantic segmentation methods so as to benefit from the high performance of the auxiliary information provided by thermal images  \cite{zhang2021abmdrnet, tu2022rgbt,tu2021multi,tu2022weakly,zhou2023lsnet,li2022rgb}.

\begin{figure}
\centering
\includegraphics[width=1\linewidth]{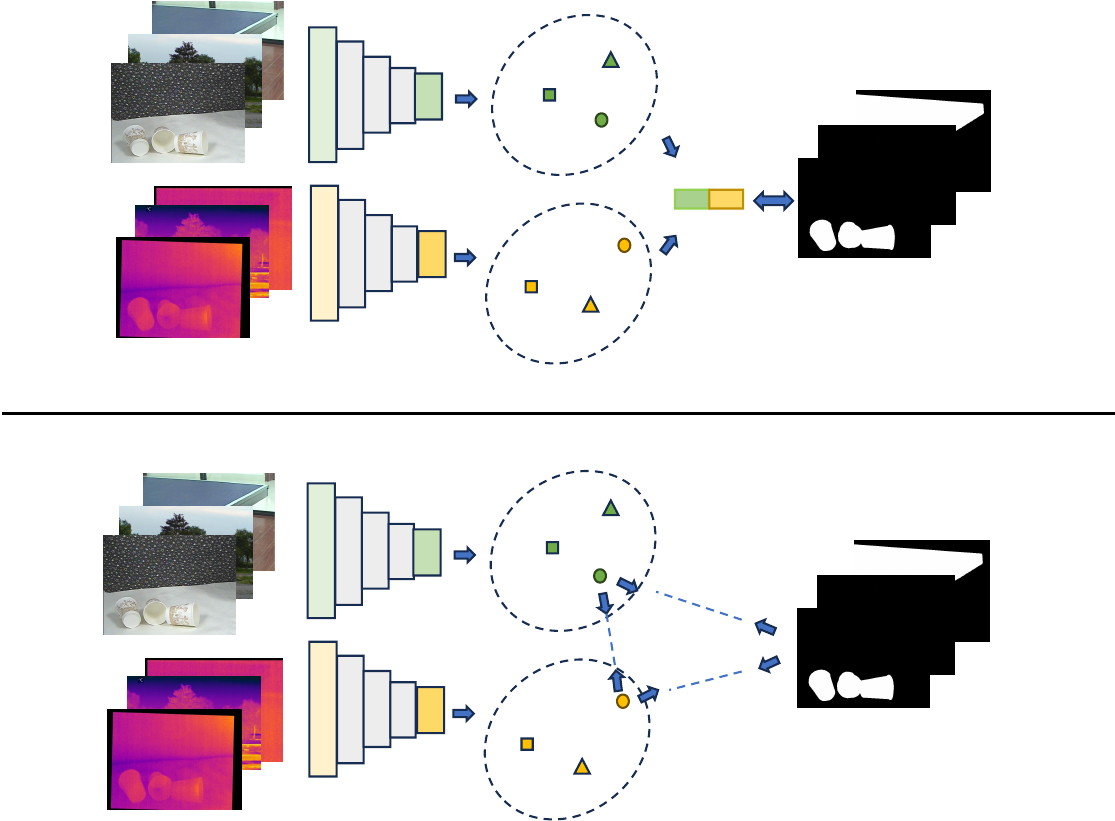}
\caption{The proposed Multi-Modal Hybrid Learning. Top: the commonly used supervised training approach. Bottom: Our proposed multi-modal hybrid learning approach jointly reduces the distance between cross-modal features, along with supervising effectively fused features.}
\label{fig:Multi-Modal Hybrid Learning}
\end{figure}

Up till now, the majority of state-of-the-art methods focus mainly on designing complex network structures to aggregate RGB and thermal features. 
LSNet \cite{zhou2023lsnet} utilizes a lightweight structure to achieve state-of-the-art performance on RGB-T segmentation. They only pay attention to fused features and neglect the semantic features of individual modalities as indicated in Figure \ref{fig:Multi-Modal Hybrid Learning}, which contain rich high-level semantic information and benefit segmentation performance \cite{liu2019poolnet,ren2020salient}.
In ABMDRNet \cite{zhang2021abmdrnet}, authors point out that the modality differences between the extracted features of individual modalities can inhibit feature fusion. In regard to this, the authors proposed a bi-directional image-to-image translation method to reduce the differences between RGB and thermal features. However, the drawbacks of this method are two-fold: i) two identical sub-networks are deployed for the modality reduction stage, leading to extra parameters and large model size; and ii) this method adopts RGB and thermal images as pseudo labels, indicating that the image pairs require perfect alignment.

To address the aforementioned drawbacks, we propose a new Multi-Modal Hybrid Learning approach that provides a new perspective and treats the thermal image as a transformation of RGB image with both supervised and self-supervised losses. Inspired by \cite{caron2018deep,grill2020bootstrap,chen2020simple}, it is designed to implicitly reduce the gap between two modalities in feature embeddings while clustering semantic features with the ground truth during the training phase. More specifically, Figure \ref{fig:Multi-Modal Hybrid Learning}
indicates that in addition to the supervision of the fused features, we supervise the semantic features from each modality and cluster cross-modal features simultaneously.

Joint training strategy has been widely used in RGB-T Segmentation \cite{zhou2022edge,zhou2023lsnet,zhang2021abmdrnet,tu2022weakly,li2022rgb}. Existing state-of-the-art methods collaboratively train the whole network with RGB and thermal images. However, in the human learning process, it is natural to learn knowledge from one area and then explore the knowledge from another area, benefiting from the previous knowledge gained. Towards this, sequential training process \cite{Pathak_2023_CVPR} for generators and classifiers in Generative Adversarial Networks (GANs) was shown to reduce knowledge gaps. In this paper, we present a novel approach to sequential training for multi-modal images, which enables robust learning of cross-modal features and results in performance gains without increasing computational overhead. Finally, by contrast with the widely used mere summation of RGB and thermal features in the spatial dimension, we present a simple but effective Hybrid Fusion Module (HFM) for the adaptive fusion of salient multi-modal features in spatial as well as channel dimensions thereby alleviating the information loss in the channel dimension.

Our main contributions are:
\begin{itemize}
    \item We propose a Multi-Modal Hybrid loss that consists of a self-supervised and a supervised loss. It implicitly reduces the gap between different modalities. The proposed training loss can effectively cluster cross-modal feature representations and further alleviate the alignment issue between image pairs.
    \item We design a novel sequential training strategy for RGB-T segmentation. This method splits the joint-training process into two stages and progressively learns RGB and thermal features. Experimental results demonstrate that sequential training improves the saliency detection performance without requiring additional data as well as any increase in parameters.
    \item A novel HFM that takes both channel and spatial information into consideration and selectively fuses cross-modal features according to RGB channel-wise weights, leading to further refining the features during features aggregation. Extensive experimental results demonstrate that the proposed methods can improve the segmentation accuracy over multiple networks and datasets.
    
\end{itemize}

\section{Related Work}
\label{sec:related}
\paragraph{RGB-T Segmentation Approaches}
\label{sec:rgbt seg approaches}
CNN-based methods for RGB-T segmentation have achieved high performance given its robustness in feature representation. EGFNet \cite{zhou2022edge}, with an edge-aware guidance fusion network, explores and embeds edge information in features fusion for RGB-T scene parsing. LASNet \cite{li2022rgb} investigates and considers the characteristics of cross-modal features at different scales. In addition to enhancing features, ABMDRnet \cite{zhang2021abmdrnet} leverages a novel adaptive-weighted bi-directional modality difference reduction network to reduce the modality differences between the extracted RGB and thermal features for better fusion. While these methods focus on feature fusion, they rely on complex network structures, adding the computational overhead \cite{zhou2022edge, li2022rgb, zhang2021abmdrnet}. In the domain of RGB-D, NANet \cite{zhang2021NonLocal} shows the effectiveness of non-local fusion of information in spatial and channel dimensions with supervised training for semantic segmentation. As thermal and RGB images are acquired using separate cameras, pixel-level alignment between these images is not achievable. DCNet \cite{tu2022weakly} focuses on the unaligned issue and proposes a deep correlation network to explore the correlations across RGB and thermal modalities for weakly alignment-free RGB-T saliency detection. LSNet \cite{zhou2023lsnet} presents a lightweight architecture along with semantic transfer learning and geometric transfer learning to enhance semantic commonality and spatial consistency. The mentioned above methods mainly leverage novel architectures to fuse multi-modal features and jointly train the RGB and thermal streams of the respective network architectures.

\paragraph{Self-Supervised Learning}
Self-supervision, as a pretext task has gained wider attention by the research community as it enables a network to learn representations without incurring the cost of annotations \cite{albelwi2022survey, liu2021self}. The representation learned using such a pre-training regime has shown to be highly effective in a variety of downstream tasks including semantic segmentation \cite{fang_selfsupervised_2022}. Furthermore, in the context of multi-modal data, \cite{valada_selfsupervised_2020} presents a segmentation framework that dynamically adapts the fusion of modality-specific features based on their saliency, instead of mere concatenation of independently learned features.

\begin{figure*}
\centering
\includegraphics[width=1\linewidth]{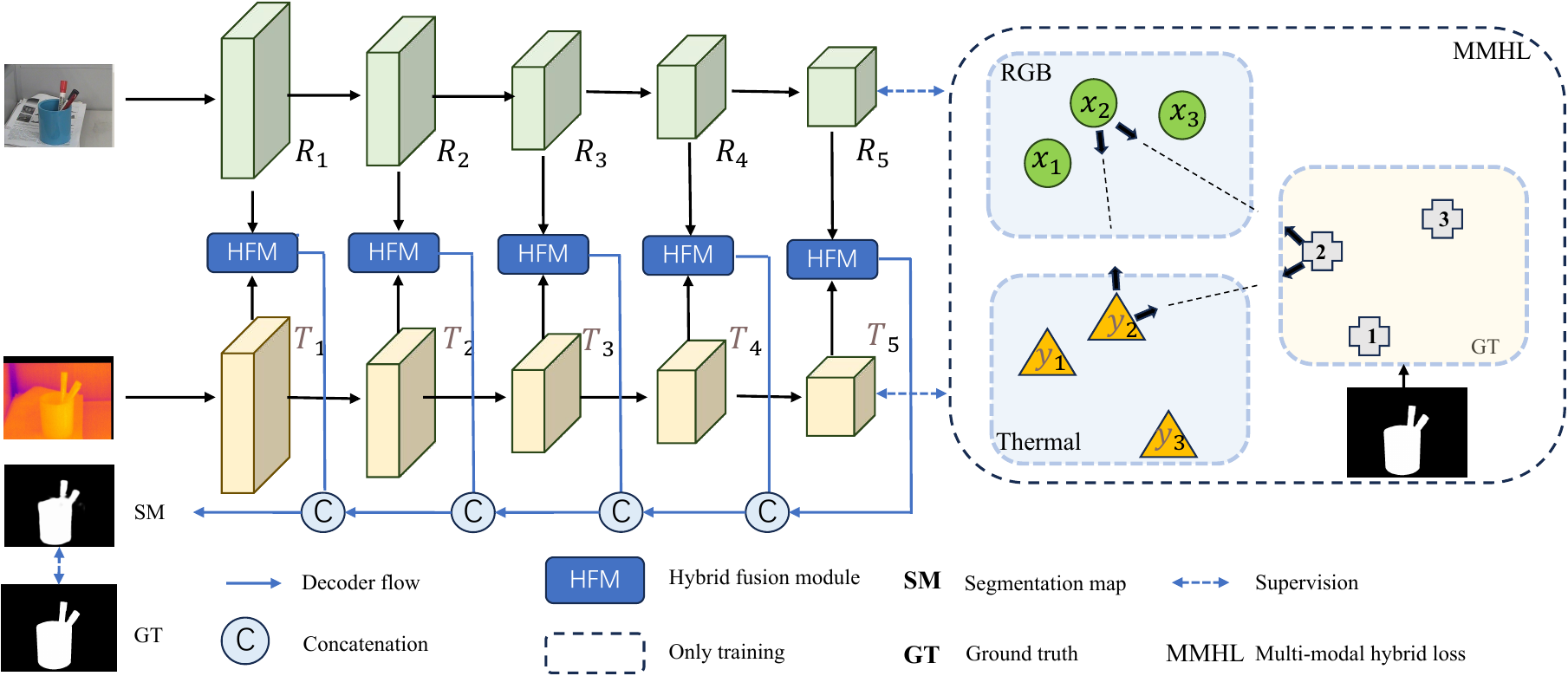}
\caption{The architecture of the proposed method. $x_{i}$ and $y_{i}$ refer to features from the $i$-th RGB and thermal images.}
\label{fig:overall achitecture}
\end{figure*}

To train such a fusion mechanism, self-supervised learning (SSL) has proven effective leading to the utilization of complementary information obtained from each modality \cite{valada_selfsupervised_2020}. Self-supervised multimodal puzzle-solving task \cite{taleb2021multimodal} combines multiple modalities at the data level to efficiently learn cross-modal complementary information and demonstrates the use of cross-modal image translation for self-supervised tasks. Among the two commonly used strategies of SSL, i.e. auxiliary pretext task \cite{zhao2022SelfSupervised} and contrastive learning \cite{caron2018deep,grill2020bootstrap,chen2020simple}, the latter has shown state-of-the-art performance for different downstream tasks \cite{albelwi2022survey}. Different from existing methods, we introduce a hybrid fusion module for the spatial and channel-level fusion of multi-modal features. For effective learning, we introduce Multi-Modal Hybrid loss which combines contrastive learning to learn cross-modal features with supervised learning to distinctly learn semantic features from each modality.


\section{Methods}
\label{sec:method}


\subsection{Overview of Framework}
We adopt two identical Res2net \cite{gao2019res2net} networks as backbones for RGB and thermal images. Learnable parameters are denoted as $\theta$ and $\xi$ for two streams. The overall architecture is shown in Figure \ref{fig:overall achitecture}. From shallow to deep layers, the RGB and thermal features are represented by $R_{i}$ and $T_{i}$ ($i \in \{1, 2, \dots, 5\}$) respectively. In order to reduce the computational overhead, two encoders share the weights and $1\times1$ convolution is employed to reduce channels for multiple levels. 

\subsection{Multi-Modal Hybrid Loss}
Multi-modal semantic features help improve segmentation performance whereas contrastive learning enables networks to learn robust representations. We propose a hybrid loss, which we refer to as Multi-Modal Hybrid Loss (MMHL), to help the network learn complementary information from RGB and thermal features and to reduce the distance between semantic features from the respective modality and the ground truth (Figure \ref{fig:overall achitecture}). 

More specifically, the MMHL consists of a self-supervised loss and a supervised loss. Inspired by \cite{grill2020bootstrap}, we first average the feature maps $R_{5}$ and $T_{5}$ on channels by global average pooling (GAP) operations:
\begin{equation}
 \centering
 S_{r} = GAP(R_{5})
\end{equation}
\begin{equation}
 \centering
 S_{t} = GAP(T_{5})
\end{equation}
We use the self-supervised loss \cite{grill2020bootstrap} between the normalized RGB and thermal feature representations:
\begin{equation}
 \centering
 L^{u}_{\theta,\xi} \stackrel{\Delta}{=} \parallel S_{r}(\theta) - S_{t}(\xi) \parallel^2_2 = 2-2 \frac{\langle S_{r}(\theta), S_{t}(\xi) \rangle}{ \parallel S_{r}(\theta) \parallel^2 \cdot \parallel S_{t}(\xi) \parallel^2}
\end{equation}
Further, we use two 1 $\times$ 1 convolution layers to decode $R_{5}$ and $T_{5}$:
\begin{equation}
    \centering
    D_{r} = conv(R_{5})
\end{equation}
\begin{equation}
    \centering
    D_{t} = conv(T_{5})
\end{equation}
The supervised loss component for two modalities is calculated using $D_{r}$, $D_{t}$ and the corresponding ground truth $G$:
\begin{equation}
    \centering
    L^{s}_{\theta,\xi} =L_{ioubce}(D_{r}(\theta),G) + L_{ioubce}(D_{t}(\xi),G)
\end{equation}
where the $L_{ioubce}$ \cite{zhou2023lsnet} indicates the binary cross-entropy and intersection-over-union (IOU) loss functions:
\begin{equation}
    \centering
    L_{ioubce} =L_{iou} +L_{bce}
\end{equation}
Finally, the proposed MMHL is the sum of the self-supervised loss and supervised loss:
\begin{equation}
\centering
\label{equation: MMHL}
L =L^{u}_{\theta,\xi} + \alpha L^{s}_{\theta,\xi}
\end{equation}
where $\alpha$ is a hyper-parameter, which is empirically set to 10 in this work.

\subsection{Sequential Training for Multi-Modal Features}
Multi-modal segmentation methods widely adopt a joint-training process \cite{zhou2022edge, zhou2023lsnet}, indicating that deep-learning models are simultaneously trained with RGB images and thermal images.

\begin{figure}[h]
\centering
\includegraphics[width=1\linewidth]{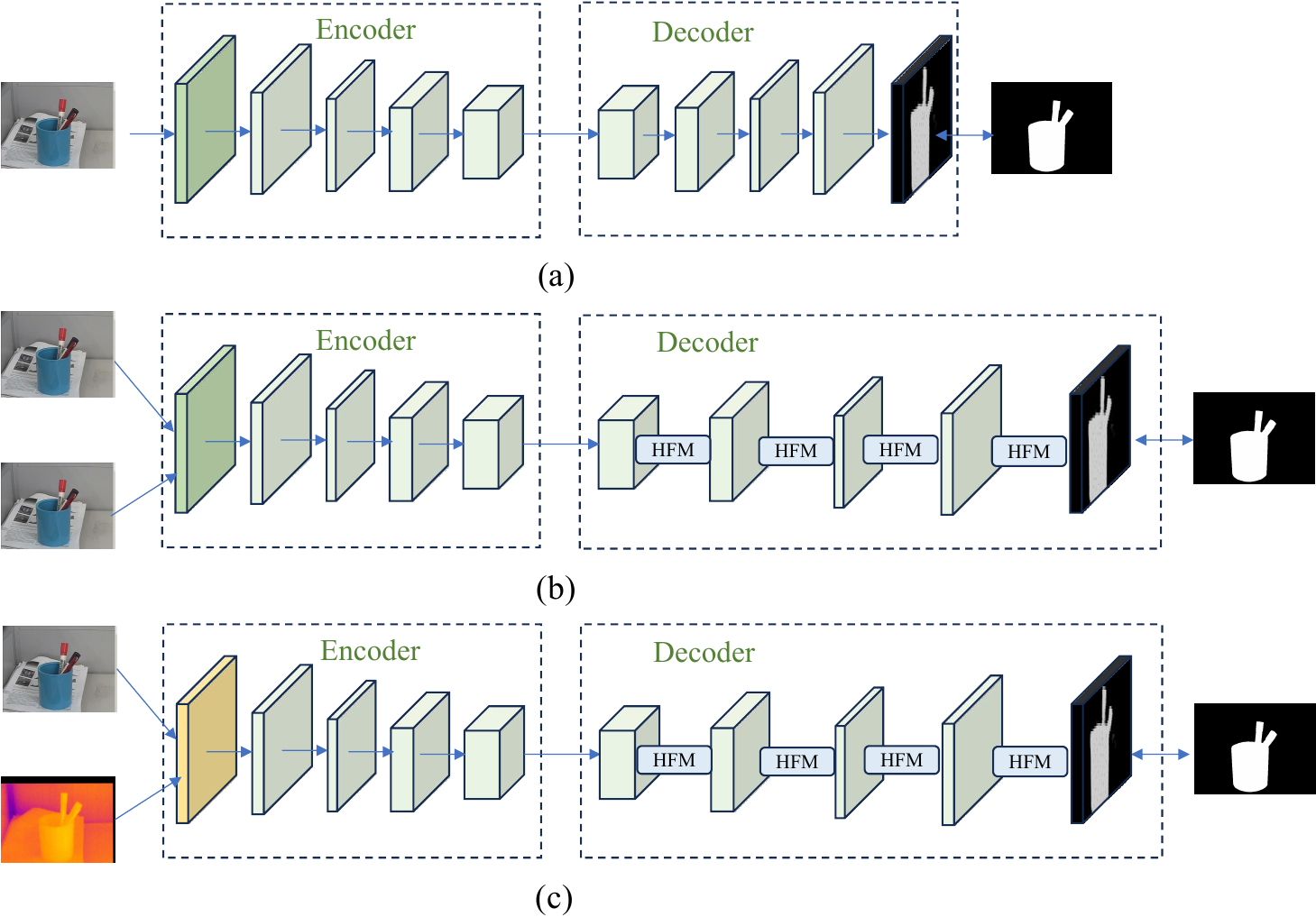}
\caption{The proposed sequential training for the training phase. (a) refers to partially sequential training, which means that we only pre-train the RGB stream. (b) refers to fully sequential training, which indicates that the whole architecture is pre-trained with RGB images. (c) is the typical training process in the second stage.}
\label{fig:sequential training}
\end{figure}

We follow the human learning process, wherein knowledge is first learned from one area and then it is extended and combined with the knowledge in another area. We call this a sequential training strategy to learn multi-modal features as shown in Figure \ref{fig:sequential training}. More specifically, instead of directly feeding multi-modal features in the network, we train the network in two stages. In the first stage, we train the encoder and decoder by solely using RGB images. Based on our approach taken for the first stage training, sequential training can be categorized into partial and full. Figure \ref{fig:sequential training}(a) refers to partial sequential training, in which only RGB encoder network stream is trained. Whereas \ref{fig:sequential training}(b) refers to fully sequential training, in which both encoder streams for RGB and thermal are pre-trained with RGB images only. In this work, we adopt the fully sequential training approach as it also enables initializing the weights for Hybrid Fusion Module during the first stage training. In the second stage, we train the network jointly with RGB images and thermal images. It is worth noting that these two training stages exploit identical hyper-parameters.
\subsection{Hybrid Fusion Module}
Previous works \cite{li2022rgb,zhou2022edge,zhou2023lsnet} mainly consider spatial information during multi-modal features fusion and they directly reduce channel numbers in order to reduce model size and parameters. To this end, inspired by the \cite{hu2018squeeze}, we propose a simple HFM to extract channel-wise features in the thermal domain by considering the RGB features. Different than \cite{zhang2021NonLocal}, HFM implements an attention mechanism \cite{oktay2018attention, hu2018squeeze} such that the RGB features are first squeezed in spatial dimension and are used to refine thermal features. These refined thermal features and RGB features are then fused as shown in the architecture of the HFM \ref{fig:hfm}. In order to take full merit of RGB features, the HFM is utilized in the sequential training process. The top structure represents the first stage, which only uses RGB features in HFM. To acquire channel-wise dependencies for RGB features, a global average pooling is applied to the spatial dimension:
\begin{figure}[htb]
\centering
\includegraphics[width=0.9\linewidth]{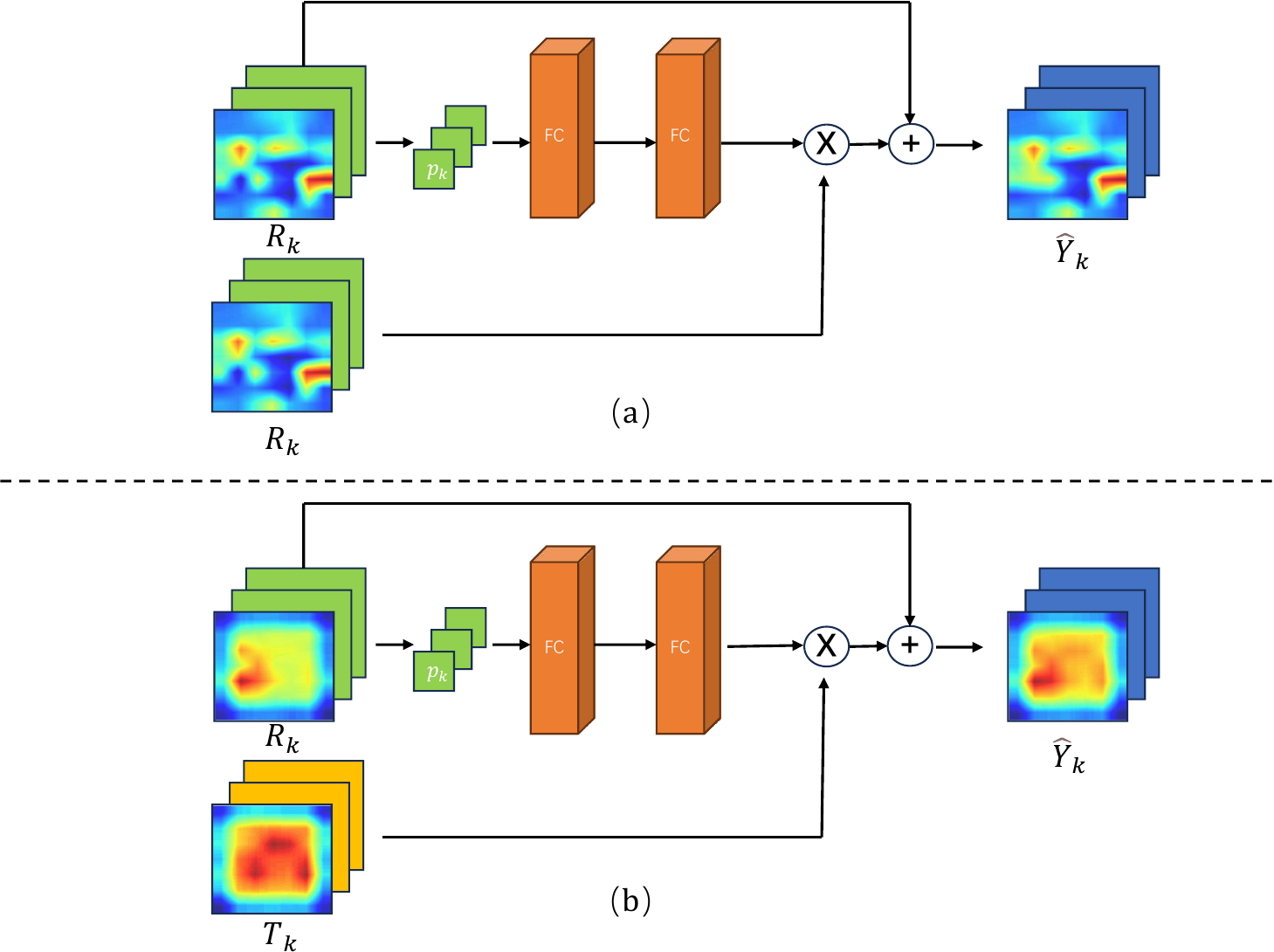}
\caption{The architecture of proposed HFM. (a) is used in the first stage of the sequential training and (b) refers to the second stage.}
\label{fig:hfm}
\end{figure}

\begin{equation}
    \centering
    p_{k} = GAP_{s}(R_{k}) = \dfrac {1}{H\times W}\sum ^{H}_{i=1}\sum ^{W}_{j=1}r_{k}\left( i,j\right)
\end{equation}
where H and W refer to the height and width in the spatial dimension respectively.
The generated channel descriptor is fed into the fully-connected layers to fully capture channel-wise dependencies for RGB modality:
\begin{equation}
    \centering
    r_{k} = F_{fc}(p_{k},\theta) = \sigma(fc_{2}(\delta(fc_{1}(p_{k},\theta_{1})),\theta_{2}))
\end{equation}
where $\sigma$ and $\delta$ refer to the Sigmoid and Rectified Linear Unit (ReLU) function. Finally, this channel-wise vector $r_{k}$ guides the RGB features $R_{k}$ at the first training stage and thermal features $T_{k}$ at the second training stage to generate the fused features $\hat{Y^{1}}$ and $\hat{Y^{2}}$:
\begin{equation}
    \centering
    \hat{Y^{1}_{k}} = r_{k}\cdot R_{k} + R_{k}
\end{equation}

\begin{equation}
    \centering
    \hat{Y^{2}_{k}} = r_{k}\cdot T_{k} + R_{k}
\end{equation}

\section{Experiments}
\label{sec:experiment}

\begin{table*}
\centering
  \caption{Performance comparisons including $F_{m}$, $S_{m}$, $E_{m}$, and $M$, with other state-of-the-art methods on three testing datasets. The best results are in \textbf{bold}.}
\label{tab:total comparisons}
\begin{tabular}{ccccc|cccc|cccc}
\toprule
\multirow{2}{*}{\textbf{Method}\vspace{-6pt}
} & \multicolumn{4}{c}{\textbf{VT821}}  & \multicolumn{4}{c}{\textbf{VT1000}}  & \multicolumn{4}{c}{\textbf{VT5000}} \\ \cmidrule{2-13} 
 & \textbf{$S_{m}$} & \textbf{$E_{m}$} & \textbf{$F_{m}$} & \textbf{$M$} & \textbf{$S_{m}$} & \textbf{$E_{m}$} & \textbf{$F_{m}$} & \textbf{$M$} & \textbf{$S_{m}$} & \textbf{$E_{m}$}& \textbf{$F_{m}$}& \textbf{$M$}\\ \hline

PoolNet & 0.751 & 0.739 & 0.578 & 0.109 & 0.834 & 0.813 & 0.714 & 0.067 & 0.769 & 0.755 & 0.588 & 0.089\\

R3Net & 0.786 & 0.809 & 0.660 & 0.073 & 0.842 & 0.859 & 0.761 & 0.055 & 0.757 & 0.790 & 0.615 & 0.083\\

CPD & 0.827 & 0.837 & 0.710 & 0.057 & 0.906 & 0.902 & 0.834 & 0.032 & 0.848 & 0.867 & 0.741 & 0.050\\
\hline
MTMR & 0.725 & 0.815 & 0.662 & 0.109 & 0.706 & 0.836 & 0.715 & 0.119 & 0.680 & 0.795 & 0.595 & 0.114\\

M3S-NIR & 0.723 & 0.859 & 0.734 & 0.140 & 0.726 & 0.827 & 0.717 & 0.145 & 0.652 & 0.780 & 0.575 & 0.168\\

SGDL & 0.765 & 0.847 & 0.731 & 0.085 & 0.787 & 0.856 & 0.764 & 0.090 & 0.750 & 0.824 & 0.672 & 0.089\\
\hline

ADF & 0.810 & 0.842 & 0.717 & 0.077 & 0.910 & 0.921 & 0.847 & 0.034 & 0.864 & 0.891 & 0.778 & 0.048\\

MIDD & 0.871 & 0.895 & 0.803 & 0.045 & 0.915 & 0.933 & 0.880 & 0.027 & 0.868 & 0.896 & 0.799 & 0.043\\

CSRNet & 0.885 & 0.908 & 0.830 & 0.038 & 0.918 & 0.925 & 0.877 & 0.024 & 0.868 & 0.905 & 0.811 & 0.042\\

DCNet & 0.877 & 0.913 & 0.822 & 0.033 & 0.923 & \textbf{0.949} & \textbf{0.902} & \textbf{0.021} & 0.872 & 0.921 & 0.819 & 0.035\\

LSNet & 0.877 & 0.911 & 0.827 & 0.033 & 0.924 & 0.936 & 0.887 & 0.022 & 0.876 & 0.916 & \textbf{0.827} & 0.036\\

Ours & \textbf{0.892} & \textbf{0.923} & \textbf{0.830} & \textbf{0.029} & \textbf{0.929} & 0.941 & 0.893 & \textbf{0.021} & \textbf{0.886} & \textbf{0.926} & 0.823 & \textbf{0.033}\\

\midrule

\end{tabular}
\end{table*}

\begin{table*}[ht]
\centering
  \caption{Performance comparisons including $F_{m}$, $S_{m}$, $E_{m}$, and $M$, with other state-of-the-art methods on the unaligned datasets \cite{tu2022weakly}. The best results are in \textbf{bold}.}
\label{tab:comparisons for unaligned}
\begin{tabular}{ccccc|cccc|cccc}
\toprule
\multirow{2}{*}{\textbf{Method}\vspace{-6pt}
} & \multicolumn{4}{c}{\textbf{Unaligned-VT821}}  & \multicolumn{4}{c}{\textbf{Unaligned-VT1000}}  & \multicolumn{4}{c}{\textbf{Unaligned-VT5000}} \\ \cmidrule{2-13} 
 & \textbf{$S_{m}$} & \textbf{$E_{m}$} & \textbf{$F_{m}$} & \textbf{$M$} & \textbf{$S_{m}$} & \textbf{$E_{m}$} & \textbf{$F_{m}$} & \textbf{$M$} & \textbf{$S_{m}$} & \textbf{$E_{m}$}& \textbf{$F_{m}$}& \textbf{$M$}\\ \hline

R3Net & 0.727 & 0.760 & 0.565 & 0.099 & 0.815 & 0.841 & 0.710 & 0.059 & 0.729 & 0.803 & 0.565 & 0.078\\

SGDL & 0.728 & 0.807 & 0.502 & 0.098 & 0.759 & 0.841 & 0.592 & 0.096 & 0.711 & 0.786 & 0.476 & 0.102\\

ADF & 0.709 & 0.727 & 0.475 & 0.157 & 0.827 & 0.826 & 0.665 & 0.088 & 0.793 & 0.816 & 0.593 & 0.088\\

MIDD & 0.840 & 0.873 & 0.707 & 0.059 & 0.896 & 0.914 & 0.814 & 0.034 & 0.844 & 0.878 & 0.719 & 0.052\\

DCNet & 0.860 & 0.908 & 0.799 & 0.036 & \textbf{0.915} & \textbf{0.943} & \textbf{0.889} & \textbf{0.023} & 0.854 & 0.908 & 0.790 & 0.041\\

Ours & \textbf{0.873} & \textbf{0.916} & \textbf{0.800} & \textbf{0.034} & 0.914 & 0.936 & 0.870 & 0.026 & \textbf{0.864} & \textbf{0.921} & \textbf{0.792} & \textbf{0.038}\\

\midrule

\end{tabular}
\end{table*}

\subsection{Setup}
\paragraph{Datasets and Evaluation Metrics}
There are three commonly used RGB-T saliency detection benchmarking datasets, namely, VT821 \cite{wang2018rgb}, VT1000 \cite{tu2019rgb} and VT5000 \cite{tu2022rgbt}, on which we performed evaluation. We follow the training and testing settings in previous work \cite{zhou2023lsnet}. The VT5000 dataset contains 5,000 pairs of RGB-T images in total, including daytime and nighttime scenes. We use 2,500 pairs of images in the training phase and the remaining images are used for testing. VT1000 and VT821 include 1,000 and 821 RGB-T images respectively and all these images are used for testing. It is worth noting that some of the RGB images have been added with noise in the VT821 dataset in order to increase the detection difficulty \cite{zhou2023lsnet}. In addition to the aligned images, we further utilize three unaligned datasets \cite{tu2022weakly}, where random spatial transformation has been applied to thermal images in the testing phase. 

We adopt five commonly-used metrics for the evaluation, i.e., precision-recall (PR), $F$-measure~\cite{achanta2009frequencybeta2}, mean absolute error (MAE), structural measure ($S$-measure)~\cite{fan2017structure}, and enhanced-alignment measure ($E$-measure)~\cite{Fan2018Enhanced}.

$F$-measure indicates the overall performance by comprehensively considering both precision and recall:

\begin{equation}
    \centering
    F_{m}=\dfrac {\left( 1+\beta ^{2}\right) \cdot \text{precision}\cdot \text{recall}}{\beta ^{2}\cdot \text{precision}+\text{recall}}
\end{equation}

Where $\beta^{2}$ is set to 0.3 as default. 

MAE averages pixel-wise absolute error between a segmentation map and its corresponding ground truth for all pixels, which can be defined by:

\begin{equation}
    \centering
    MAE = \dfrac {1}{W\times H}\sum ^{W}_{x=1}\sum ^{H}_{y=1}\left| S\left( x,y\right) -G\left( x,y\right) \right|
    \label{equation-mae-measure}
\end{equation}
where $W$ denotes the width and $H$ denotes the height of the prediction, $S$ denotes the segmentation map, which is the model's output, and $G$ represents the ground truth map.

$S$-measure captures structural information and assesses the structural similarity between regional perception ($S_{r}$) and object perception ($S_{o}$). Thus, $S_{\alpha}$ can be defined by

\begin{equation}
    \centering
    S_{m} = \alpha * S_{r} + (1-\alpha) * S_{o}
    \label{equation-s-measure}
\end{equation}

Where $\alpha$ is empirically set to 0.5.

$E$-measure captures image-level statistics and their local pixel-matching information simultaneously.
\begin{equation}
    \centering
    E_{m} = \dfrac {1}{W\times H}\sum ^{W}_{x=1}\sum ^{H}_{y=1}\theta_{FM}(x,y)
    \label{equation-e-measure}
\end{equation}

Where $\theta_{FM}$ represents the enhanced-alignment matrix

\paragraph{Implementation Details}
Our model is implemented using Pytorch Toolbox and trained on an NVIDIA GeForce RTX 3090 GPU with a mini-batch size of 10. We follow the initial learning rate as used in \cite{zhou2023lsnet} and resize the training and testing images to $224\times224$. We adopt a 0.0005 weight decay for the Stochastic Gradient Descent (SGD) with a momentum of 0.9.

\subsection{Comparison with the state-of-the-arts}
We compare our model with 11 state-of-the-art methods, including three RGB-based methods CPD~\cite{wu2019cascaded}, R3Net~\cite{deng2018r3net}, PoolNet~\cite{liu2019poolnet}, and traditional methods MTMR~\cite{wang2018rgb},SGDL~\cite{tu2019rgb}, M3S-NIR~\cite{tu2019m3s}, and the latest methods ADF~\cite{tu2022rgbt}, CSRNet~\cite{huo2021csrnet}, MIDD~\cite{tu2021multi}, DCNet~\cite{tu2022weakly} and LSNet~\cite{zhou2023lsnet}. For fair comparisons, we utilize the segmentation maps provided by the authors and the pre-calculated evaluation results directly.

\begin{figure*}
\centering
\includegraphics[width=\linewidth]{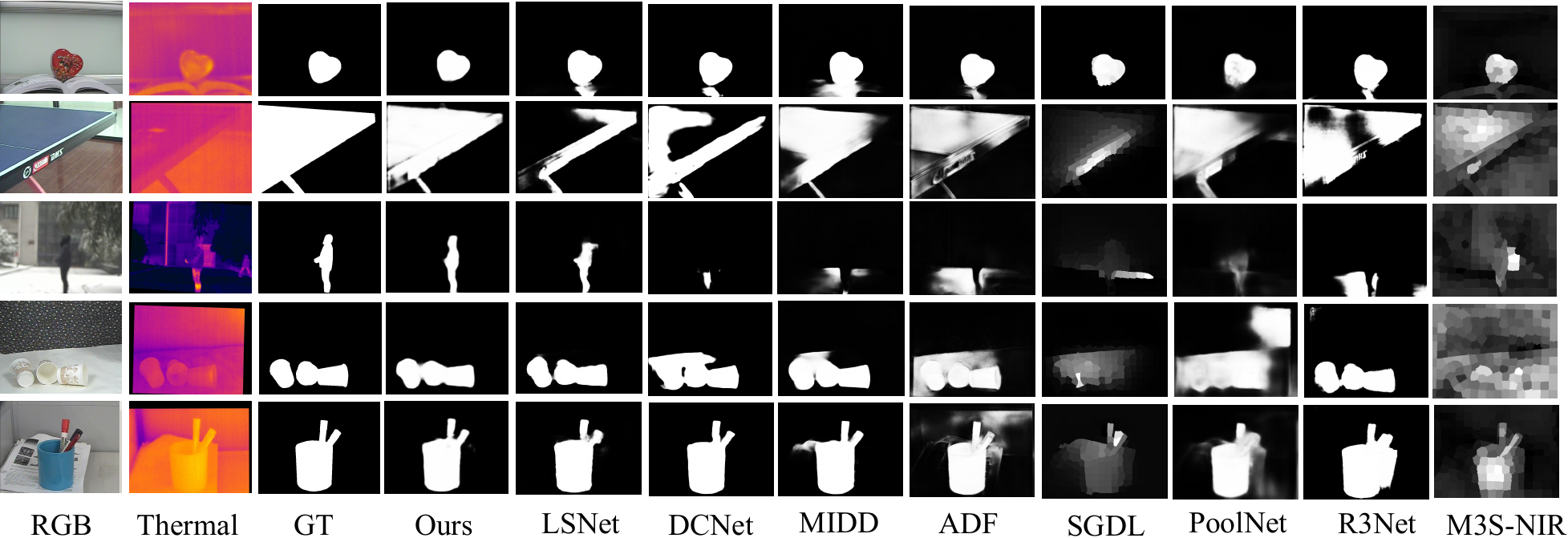}
\caption{Total comparisons with other state-of-the-art methods.}
\label{fig:total comparisons}
\end{figure*}

\begin{figure*}[ht]
 \minipage{0.32\textwidth}
 \centering
 \includegraphics[width=0.9\linewidth]{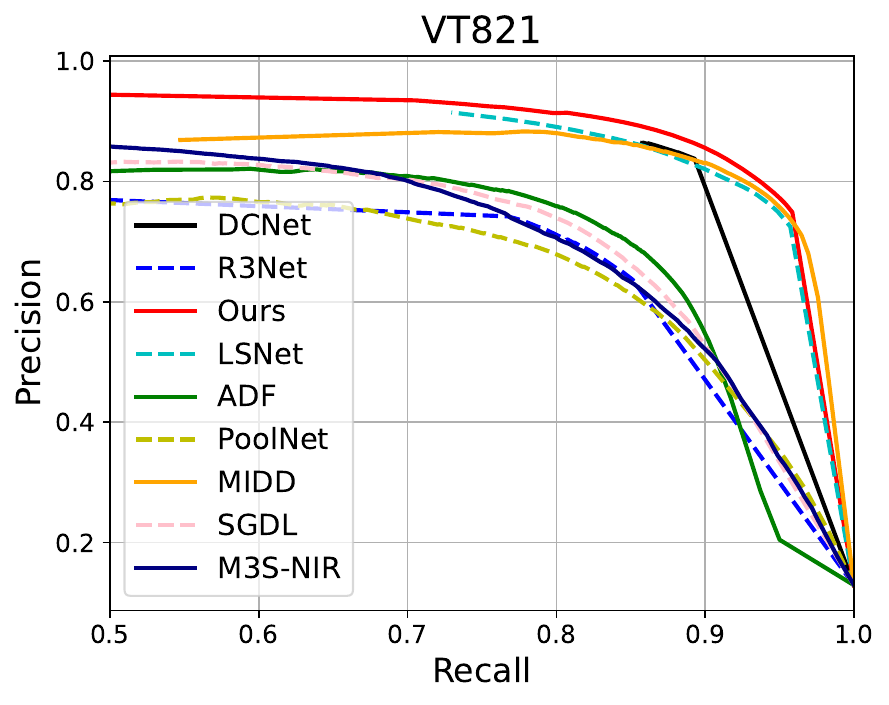}
 (\textbf{a}) 
 \endminipage\hfill
 \minipage{0.32\textwidth}%
 \centering
 \includegraphics[width=0.9\linewidth]{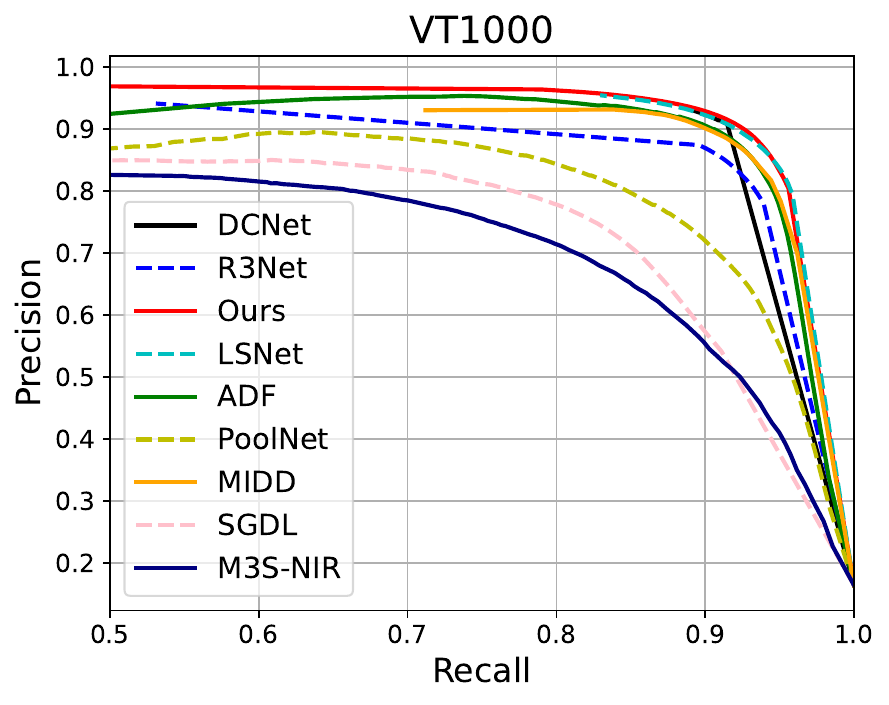}
 (\textbf{b}) 
 \endminipage\hfill
 \minipage{0.32\textwidth}
 \centering
 \includegraphics[width=0.9\linewidth]{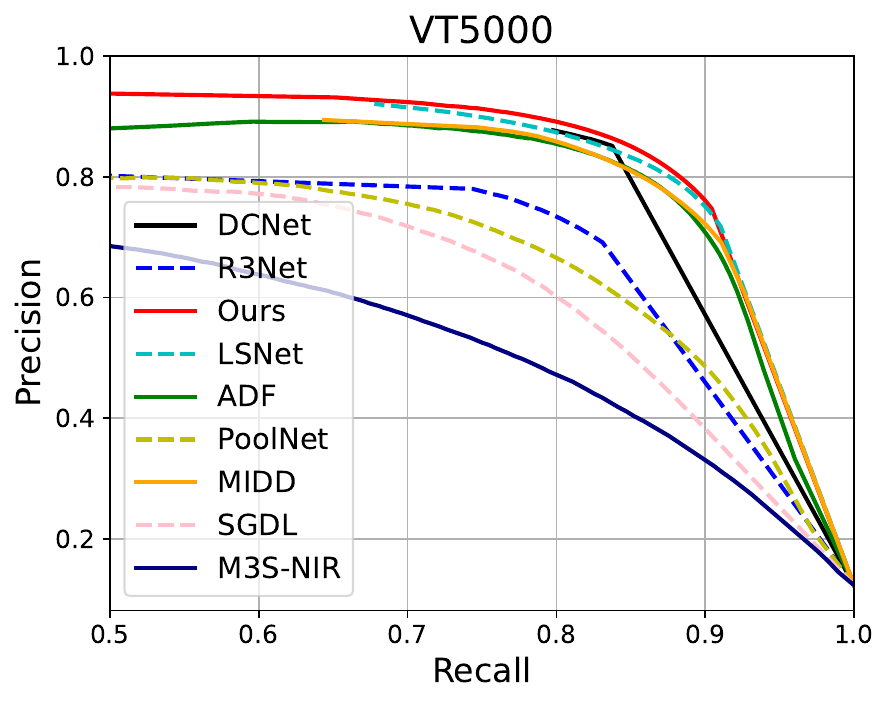}
 (\textbf{c}) 
 \endminipage\hfill
 
\caption{PR curves across three benchmarks}
 \label{fig:pr curves}
\end{figure*}

\noindent{\textbf{Quantitive Comparisons}} Table \ref{tab:total comparisons} and Figure \ref{fig:pr curves} show
the overall quantitative performance of the methods. First of all, it can be clearly observed that there is a large gap between the traditional methods, namely, MTMR, M3S-NIR, SGDL, and deep-learning methods.
However, RGB-based approaches such as PoolNet were not able to achieve accurate results due to the lack of thermal information. Compared with other approaches, our method surpasses all of them in terms of $S_{m}$ and MAE across three testing datasets. Especially, our method outperforms all others by a large margin in four metrics on the VT821, which has added noise on some RGB images in order to increase the detection difficulty. Figure \ref{fig:pr curves} shows the overall comparison results on PR curves and our method, represented by the red line, is on top of the other state-of-the-art methods. Additionally, to verify the alignment ability, we further conduct experiments on three unaligned datasets. Table \ref{tab:comparisons for unaligned} shows that our model can achieve accurate results on unaligned RGB-T image pairs, especially for the unaligned VT821 and VT5000, where our method obtains the best results over four evaluation metrics.

\noindent{\textbf{Qualititive Comparisons}}
Figure \ref{fig:total comparisons} illustrates the visual comparisons for all state-of-the-art approaches. It can be clearly observed that the segmentation maps generated by our model are close to the ground truth. Concretely, the first two rows show that our model can effectively fuse cross-modal features with fewer false positives and negatives for simple scenes. The RGB image in the third row has been added noise, under this circumstance, our model still generates a satisfactory map compared with other methods. In addition, RGB and thermal images are not perfectly aligned in reality. Maps in the last two rows illustrate that our model is able to tackle the unaligned issue in complex scenarios (e.g. wherein separate RGB and thermal cameras are installed).

\subsection{Ablation Study}
To evaluate the effectiveness of the proposed methods, we construct a robust baseline through two Res2nets, which share weights during training and testing to reduce the number of parameters. The RGB and thermal features are directly combined by simple addition. Then, we progressively concatenate multi-scale features from deep to shallow layers. We present the results of the ablation study to highlight the contribution of each component including the multi-modal features, Multi-Modal Hybrid Loss (MMHL), Hybrid Fusion Module (HFM) and sequential training approach. Results in Table \ref{table:ablation} shows increasing performance with the addition of each of the mentioned components, with the best performance resulting from the combination of all.

\begin{table}[htb]
\caption{Ablation analyses on VT821, VT1000 and VT5000. Each component is added on top of the previous component.}
\label{table:ablation}
\resizebox{\columnwidth}{!}{
\begin{tabular}{cc|ccccc}
\hline
\hline
\multirow{2}{*}{} & Metric & RGB & RGB-T  & +MMHL  &+HFM  & +Sequential  \\

                         \hline
\multirow{4}{*}{{VT821}~}    & $F_{m}\uparrow$  & 0.758 & 0.792 & 0.814 &0.821  & \textbf{0.830} \\
                         & $S_{m}\uparrow$  & 0.856 & 0.878 & 0.887 & 0.890  & \textbf{0.892} \\
                         & $E_{m}\uparrow$  & 0.884 & 0.893 & 0.909 & 0.920  & \textbf{0.923} \\
                         & $M\,\downarrow$  & 0.044 & 0.037 & 0.033 & 0.031 & \textbf{0.029} \\ \hline

\multirow{4}{*}{{VT1000}~}    & $F_{m}\uparrow$ & 0.869 & 0.877 & 0.883 &0.886  & \textbf{0.893} \\
                         & $S_{m}\uparrow$  & 0.918 & 0.924 & 0.926 &0.927 & \textbf{0.929} \\
                         & $E_{m}\uparrow$  & 0.923 & 0.932 & 0.932 &0.938  &  \textbf{0.941} \\
                         & $M\,\downarrow$  & 0.026 & 0.023 & 0.023 &0.023 & \textbf{0.021} \\ \hline
\multirow{4}{*}{{VT5000}~}    & $F_{m}\uparrow$ & 0.789 & 0.793 & 0.814 &0.813  & \textbf{0.823} \\
                     & $S_{m}\uparrow$ & 0.869 & 0.878 & 0.886 & 0.882  & \textbf{0.886} \\
                         & $E_{m}\uparrow$  &  0.906 & 0.902 & 0.917 & 0.921 & \textbf{0.926} \\
                         & $M\,\downarrow$ & 0.042 & 0.037 & 0.034 &0.035 & \textbf{0.033} \\ \hline

\hline

\end{tabular}
}
\end{table}

\begin{figure}[ht]
\centering
\includegraphics[width=1\linewidth]{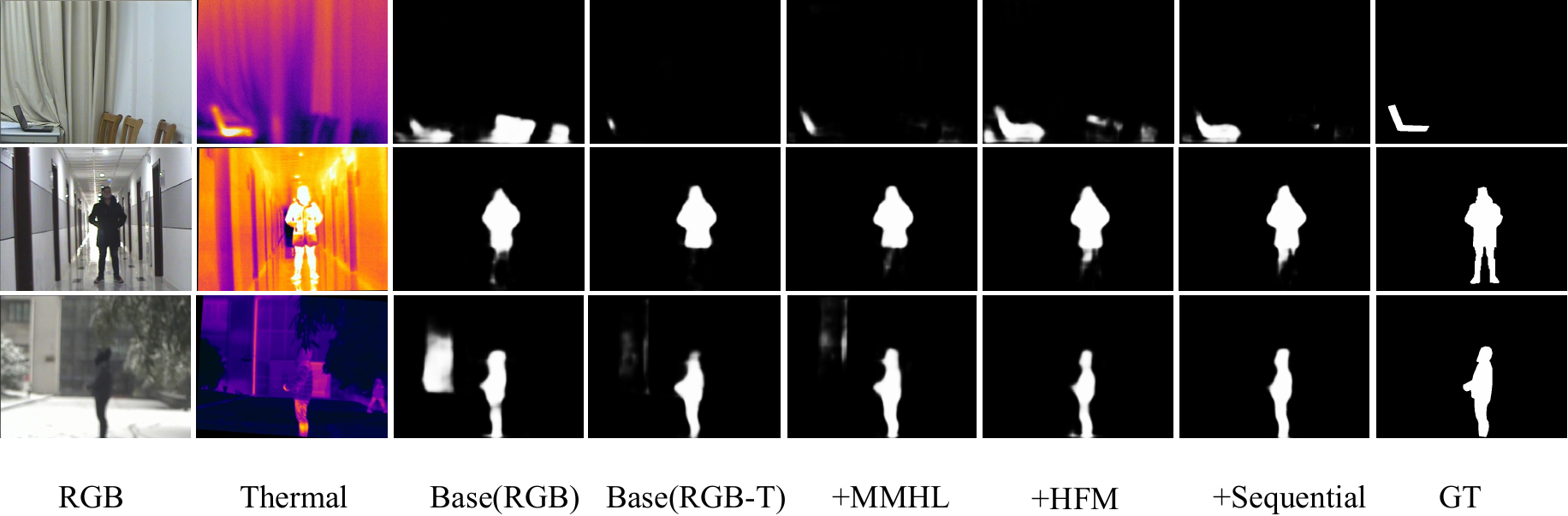}
\caption{Qualitative results of ablation study highlighting the effectiveness of the proposed components. Each component is added on top of the previous component.}
\label{fig:ablation}
\end{figure}






\subsubsection{Multi-Modal Hybrid Loss}

\begin{figure}[hb]
 \minipage{0.5\linewidth}
 \centering
 \includegraphics[width=\linewidth]{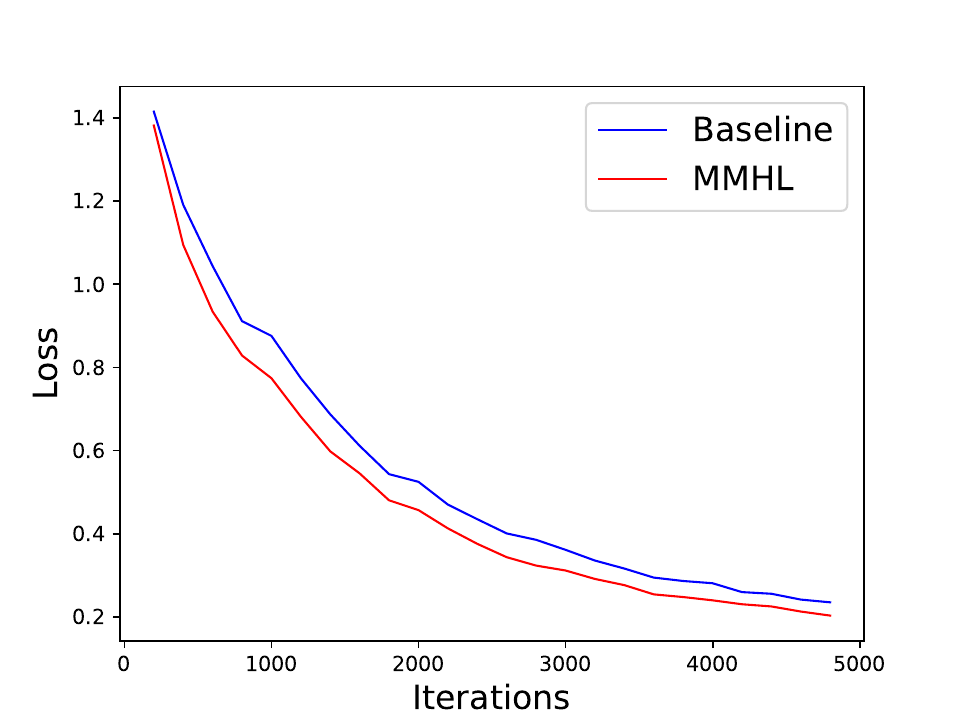}
 (\textbf{a}) 
 \endminipage\hfill
 \minipage{0.5\linewidth}%
 \centering
 \includegraphics[width=\linewidth]{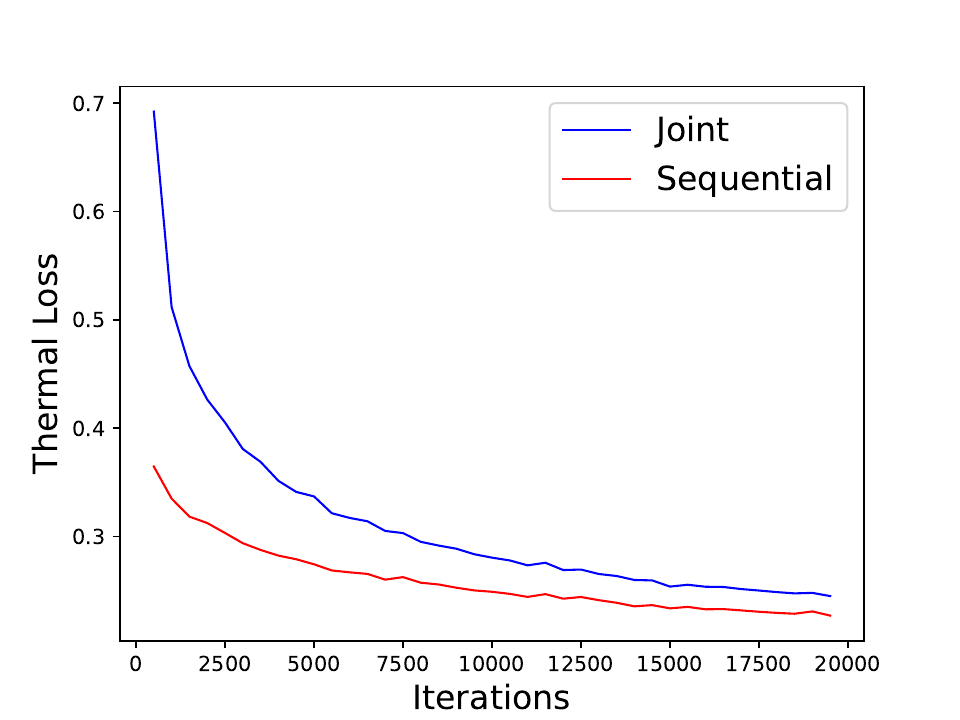}
 (\textbf{b}) 
 \endminipage\hfill
\caption{(a) refers to the losses of the final decoder in Figure \ref{fig:overall achitecture}. (b) refers to the losses only from the thermal decoder. }
 \label{fig:hssl and thermal loss}
\end{figure}

The proposed MMHL can steadily increase the detection accuracy on the baseline network across three datasets. Figure \ref{fig:ablation} illustrates that compared to the baseline, the MMHL can extract more features from thermal modality due to the contrastive learning between cross-modal features. We show the losses during the training phase in Figure \ref{fig:hssl and thermal loss} (a). Equipped with the MMHL, the model converges faster and approaches lower loss compared to the baseline. In other words, the proposed MMHL reduces the gap between RGB and thermal representations and helps the model to effectively learn cross-modal features. We also implement the LSNet baseline and employ our loss on it. Table \ref{table:hssl on lsnet} indicates that the proposed loss can improve the performance on different architectures. 

\begin{table}[ht]
\caption{Ablation study with the proposed MMHL on LSNet Network.}
\label{table:hssl on lsnet}
\centering
\resizebox{0.75\columnwidth}{!}{
\begin{tabular}{cc|cc}
\hline
\hline
\multirow{2}{*}{} & Metric   & LSNet(base)  &+MMHL    \\

                         \hline
\multirow{4}{*}{{VT821}~}    & $F_{m}\uparrow$  & 0.742 &0.780    \\
                         & $S_{m}\uparrow$   & 0.849 & 0.870    \\
                         & $E_{m}\uparrow$   & 0.881 & 0.901    \\
                         & $M\,\downarrow$  & 0.045 & 0.036   \\ \hline

\multirow{4}{*}{{VT5000}~}    & $F_{m}\uparrow$  & 0.761 &0.781    \\
                         & $S_{m}\uparrow$   & 0.857 &0.865  \\
                         & $E_{m}\uparrow$   & 0.893 &0.908     \\
                         & $M\,\downarrow$   & 0.044 &0.039   \\ \hline

\hline

\end{tabular}
}
\end{table}

\subsubsection{Sequential Training}
In this paper, we adopt the fully sequential training (Figure \ref{fig:sequential training} (b)), which indicates that the whole architecture including the HFM is trained in the first stage by RGB features. Table \ref{table:ablation} shows that this sequential training strategy can effectively improve detection performance without requiring extra parameters. We further conduct experiments on partially sequential training (Figure \ref{fig:sequential training} (a)), where only the RGB encoder and decoder are trained in the first stage. Table \ref{table:sequential} demonstrates that even partially sequential training can boost accuracy and reduce the error rate. 
In order to investigate thermal features in sequential training, we show the losses from the thermal decoder with/without sequential training in Figure \ref{fig:hssl and thermal loss} (b). It can be apparently observed that sequential training can help the model start from a small loss due to the learned RGB features and the training loss keeps lower than the loss trained in a typical way.

\begin{table}[ht]
\caption{Partially sequential training method. We only train the RGB encoder and decoder in the first stage as shown in Figure \ref{fig:sequential training} (a).}
\label{table:sequential}
\centering
\resizebox{0.75\columnwidth}{!}{
\begin{tabular}{cc|cc}
\hline
\hline
\multirow{2}{*}{} & Metric   & +MMHL  &+Sequential    \\

                         \hline
\multirow{4}{*}{{VT821}~}    & $F_{m}\uparrow$  & 0.814 &0.820    \\
                         & $S_{m}\uparrow$   & 0.887 & 0.887    \\
                         & $E_{m}\uparrow$   & 0.909 & 0.914    \\
                         & $M\,\downarrow$  & 0.033 & 0.032   \\ \hline

\multirow{4}{*}{{VT1000}~}    & $F_{m}\uparrow$  & 0.883 &0.890    \\
                         & $S_{m}\uparrow$   & 0.926 &0.928  \\
                         & $E_{m}\uparrow$   & 0.932 &0.937     \\
                         & $M\,\downarrow$   & 0.023 &0.022   \\ \hline

\hline

\end{tabular}
}
\end{table}

\subsubsection{Hybrid Fusion Module}
Table \ref{table:ablation} shows that instead of simply adding cross-modal features, the HFM investigates the channel-wise mutual information for thermal features and alleviates the loss of channel features for the thermal modality, leading to the improvement of overall performance on $E_{m}$, especially on VT821, where the scenarios are more complicated. Figure \ref{fig:ablation} also illustrates that with the addition of the HFM, the model can obtain more true positives (column 6) meanwhile mitigating the false positives (row 3).

\section{Conclusions}
In this paper, we leverage novel self-supervised learning and sequential training for multi-modal RGB-T segmentation. Considering the differences between RGB and thermal features, we propose to treat the thermal images as a transformation of RGB images and introduce self-supervised learning to cluster the feature embeddings. Moreover, to take full advantage of cross-modal features, we follow the human learning process and propose a sequential training strategy to split the training phase into two stages. We further design a novel module to consider both channel and spatial information instead of naively concatenating RGB and thermal features. Experimental results demonstrate the superiority and effectiveness of the proposed methods on three aligned benchmarking datasets, as well as their unaligned versions.
\newpage
\section{Supplementary}

\subsection{Multi-Modal Hybrid Loss}

\begin{figure}[ht]
\centering
\includegraphics[width=\linewidth]{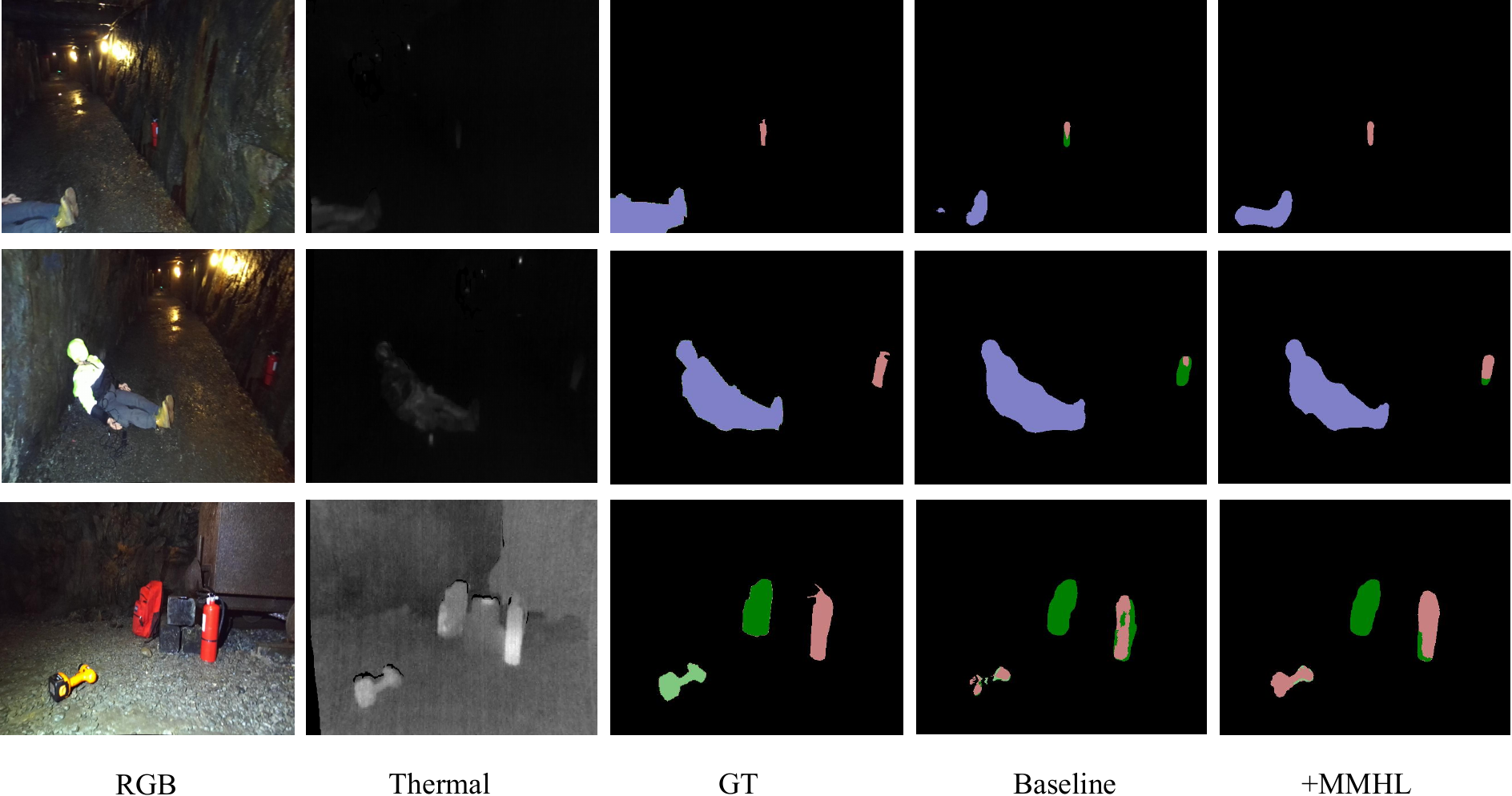}
\caption{Qualitative results of ablation study on PST900 highlighting the effectiveness of the proposed MMHL.}
\label{fig:pst900}
\end{figure}

To further demonstrate the applicability of the proposed MMHL, we train a different network using the proposed MMHL for a different task. More specifically, we change the backbone in Figure \ref{fig:overall achitecture} to VGG16 \cite{simonyan2014vgg} and to evaluate the effectiveness only of MMHL, we do not add HFM and use the commonly used joint training strategy instead of the sequential training as proposed in this work. We exploit PST900 dataset  \cite{shivakumar2020pst900} for semantic segmentation task. This dataset has 894 synchronized and calibrated RGB and Thermal image pairs across four distinct classes. We follow the evaluation metrics as used in PST900 \cite{shivakumar2020pst900}, i.e. mean IOU (mIoU) to evaluate the segmentation performance. Table \ref{table:pst900} and Figure \ref{fig:pst900} demonstrate effectiveness of the proposed MMHL when applied to different architectures as well as tasks.
\begin{table}[ht]
\caption{Ablation study with the proposed MMHL on the ResNet50 backbone.}
\label{table:hssl on resnet50}
\centering
\resizebox{0.75\columnwidth}{!}{
\begin{tabular}{cc|cc}
\hline
\hline
\multirow{2}{*}{} & Metric   & baseline  &+MMHL    \\

                         \hline
\multirow{4}{*}{{VT821}~}    & $F_{m}\uparrow$  & 0.770 &0.784    \\
                         & $S_{m}\uparrow$   & 0.867 & 0.872    \\
                         & $E_{m}\uparrow$   & 0.884 & 0.891    \\
                         & $M\,\downarrow$  & 0.040 & 0.038   \\ \hline

\multirow{4}{*}{{VT1000}~}    & $F_{m}\uparrow$  & 0.861 &0.867    \\
                         & $S_{m}\uparrow$   & 0.915 &0.917  \\
                         & $E_{m}\uparrow$   & 0.920 &0.927     \\
                         & $M\,\downarrow$   & 0.027 &0.025   \\ \hline

\hline

\end{tabular}
}
\end{table}

\begin{table}[hb]
\caption{Alation study on PST900.}
\label{table:pst900}
\centering
\resizebox{0.9\columnwidth}{!}{
\begin{tabular}{cc|cccc}
\hline
\hline
\multirow{2}{*}{} & Metric &2VGG16  & 2VGG16+MMHL      \\

                         \hline
\multirow{1}{*}{{PST900}~}    & $mIOU\uparrow$  & 0.500 & \textbf{0.558} \\ \hline

\hline

\end{tabular}
}
\end{table}

\subsection{Sequential Training}

We conduct experiments on partially sequential training (Figure \ref{fig:sequential training} (a)), where only the RGB encoder and decoder are trained in the first stage. Table \ref{table: supp sequential} demonstrates that even partially sequential training can boost accuracy and the final results are competitive with that obtained using fully sequential training as highlighted in the table \ref{table:ablation}.

\begin{table}[hb]
\caption{Partially sequential training method.}
\label{table: supp sequential}

\resizebox{\columnwidth}{!}{
\begin{tabular}{cc|ccccc}
\hline
\hline
\multirow{2}{*}{} & Metric & RGB & RGB-T  & +HSSL  &+Sequential  & +HFM  \\

                         \hline
\multirow{4}{*}{{VT821}~}    & $F_{m}\uparrow$  & 0.758 & 0.792 & 0.814 &0.820  & \textbf{0.829} \\
                         & $S_{m}\uparrow$  & 0.856 & 0.878 & 0.887 & 0.887  & \textbf{0.893} \\
                         & $E_{m}\uparrow$  & 0.884 & 0.893 & 0.909 & 0.914  & \textbf{0.917} \\
                         & $M\,\downarrow$  & 0.044 & 0.037 & 0.033 & 0.032 & \textbf{0.029} \\ \hline

\multirow{4}{*}{{VT1000}~}    & $F_{m}\uparrow$ & 0.869 & 0.877 & 0.883 &0.890  & \textbf{0.893} \\
                         & $S_{m}\uparrow$  & 0.918 & 0.924 & 0.926 &0.928 & \textbf{0.929} \\
                         & $E_{m}\uparrow$  & 0.923 & 0.932 & 0.932 &0.937  &  \textbf{0.939} \\
                         & $M\,\downarrow$  & 0.026 & 0.023 & 0.023 &0.022 & \textbf{0.021} \\ \hline

\hline

\end{tabular}
}
\end{table}

\begin{figure}[hb]
 \minipage{0.5\linewidth}
 \centering
 \includegraphics[width=\linewidth]{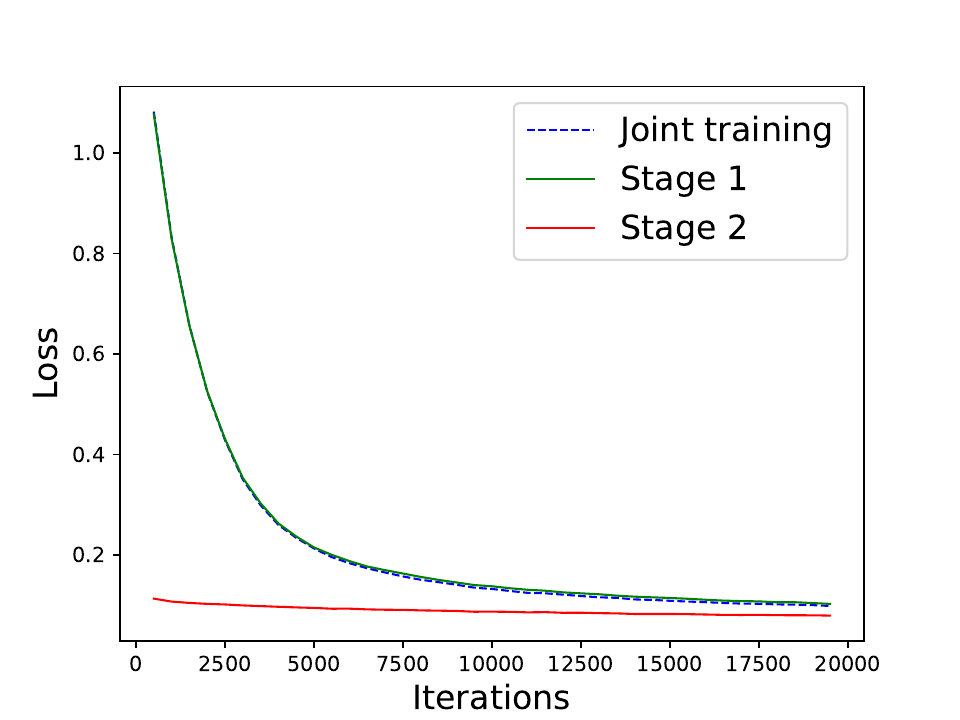}
 (\textbf{a}) 
 \endminipage\hfill
 \minipage{0.5\linewidth}%
 \centering
 \includegraphics[width=\linewidth]{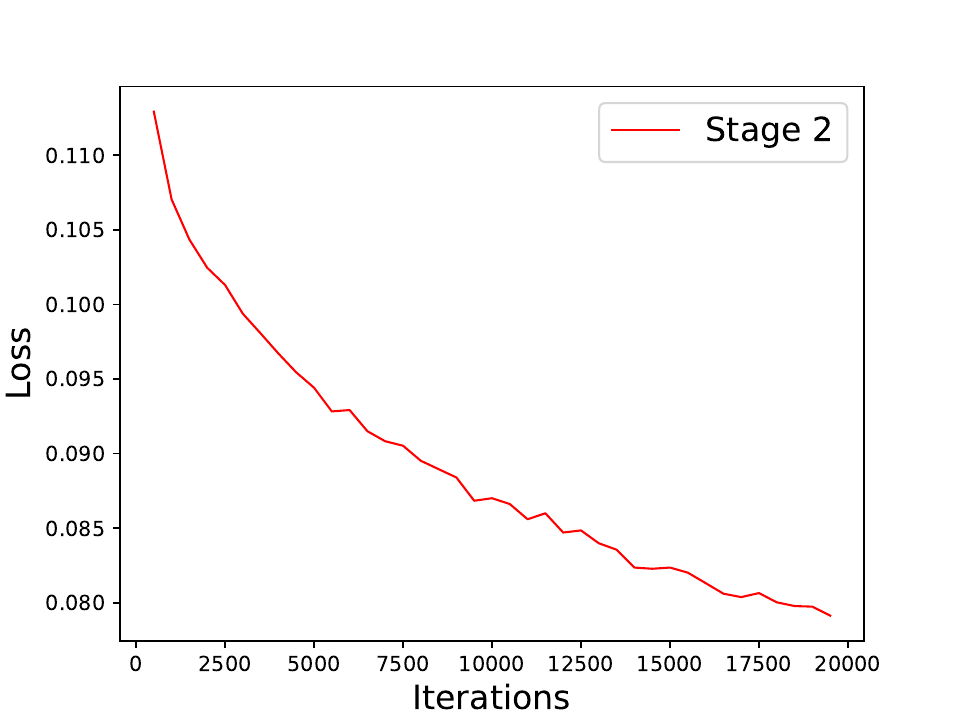}
 (\textbf{b}) 
 \endminipage\hfill

\caption{Loss curves for the sequential training. 
}
 \label{fig:supp loss}
\end{figure}

We visualize the training process with different training strategies as shown in Figure \ref{fig:supp loss} (a). More specifically, the blue curve refers to the model being trained in a typical way with both RGB and thermal images. The green curve refers to the first stage of fully sequential training with only RGB images. It can be observed that the model converges similarly. In other words, the joint training strategy cannot make full use of thermal information. Compared with it, our method splits the training into two stages and the final loss further decreases (red curve in (a) and (b)).

We further train our model with the joint training strategy and keep the same training iterations with the two-stage sequential training. In order to avoid the overfitting issue on similar image pairs, we adopt the unaligned VT821 and an RGB-D testing dataset SIP \cite{fan2020sip}, which contains 921 pairs of RGB and Depth images. Table \ref{table:joint and sequential} demonstrates that the model has a more robust generalization ability by training with the proposed sequential strategy.

\begin{table}[ht]
\caption{Quantitive comparisons between the joint training and sequential training on different testing datasets. SIP \cite{fan2020sip} is commonly used in RGB-Depth segmentation.}
\label{table:joint and sequential}
\centering
\resizebox{0.9\columnwidth}{!}{
\begin{tabular}{cc|cc}
\hline
\hline
\multirow{2}{*}{} & Metric   & Joint  & Sequential    \\

                         \hline
\multirow{4}{*}{{Unaligned-VT821}~}    & $F_{m}\uparrow$  & 0.792 &\textbf{0.800}    \\
                         & $S_{m}\uparrow$   & 0.868 & \textbf{0.873}    \\
                         & $E_{m}\uparrow$   & 0.916 & 0.916    \\
                         & $M\,\downarrow$  & 0.036 & \textbf{0.034}   \\ \hline

\multirow{4}{*}{{SIP}~}    & $F_{m}\uparrow$  & 0.784 &\textbf{0.796}    \\
                         & $S_{m}\uparrow$   & 0.840 &\textbf{0.848}  \\
                         & $E_{m}\uparrow$   & 0.896 &\textbf{0.901}     \\
                         & $M\,\downarrow$   & 0.071 &\textbf{0.067}   \\ \hline

\hline

\end{tabular}
}
\end{table}

\subsection{The Training Process}

Our final training process is illustrated in algorithm \ref{algo:fully}. The difference between partially \ref{algo:partially} and fully sequential training is the training model. For partially sequential training, we only train the RGB stream with RGB images. For fully sequential training, we train the whole network including the HFM. 
It is worth noting that the first stage of sequential training adopts identical training parameters to the second stage.

\begin{algorithm}
	\caption{ Fully Sequential training process.
	}
	\begin{algorithmic}[1]
            \STATE \textbf{Stage 1}: Training the whole model
		
            \STATE \textbf{Input}: RGB maps
            \STATE $\theta$ = $argmin_{\theta}L_{ioubce}(R,GT)$;
		\STATE \textbf{Stage 2}: Training the whole model
            \STATE \textbf{Input}: RGB and thermal maps
            \STATE $\theta,\xi$ = $argmin_{\theta,\xi}L(R,T,GT)$;

	\end{algorithmic}
	\label{algo:fully}
\end{algorithm}

\begin{algorithm}
	\caption{ Partially Sequential training process.
	}
	\begin{algorithmic}[1]
		\STATE \textbf{Stage 1}: Training the RGB model
		
            \STATE \textbf{Input}: RGB maps
            \STATE $\theta$ = $argmin_{\theta}L_{ioubce}(R,GT)$;
		\STATE \textbf{Stage 2}: Training the whole model
            \STATE \textbf{Input}: RGB and thermal maps
            \STATE $\theta,\xi$ = $argmin_{\theta,\xi}L(R,T,GT)$;

	\end{algorithmic}
	\label{algo:partially}
\end{algorithm}

\subsection{Hyperparameter Optimization}

In order to find optimal $\alpha$ for our MMHL (see Equation \ref{equation: MMHL}), we carried out hyperparameter optimization. We investigated the performance of our loss on the VT821 with different hyperparameter settings as presented in Tables \ref{table:hyper for mmhl}. Results show that the most accurate segmentation is achieved with $\alpha=10$.

\begin{table}[ht]
\caption{Different $\alpha$ for the MMHL.}
\label{table:hyper for mmhl}
\centering
\resizebox{0.85\columnwidth}{!}{
\begin{tabular}{cc|cccc}
\hline
\hline
\multirow{2}{*}{} & Metric &5  & 10  &15    \\

                         \hline
\multirow{4}{*}{{VT821}~}    & $F_{m}\uparrow$  & 0.811 & 0.814 &0.812   \\
                         & $S_{m}\uparrow$   & 0.883 & 0.887 & 0.885    \\
                         & $E_{m}\uparrow$   & 0.908 & 0.909 & 0.911   \\
                         & $M\,\downarrow$  & 0.033 & 0.033  & 0.033 \\ \hline

\hline

\end{tabular}
}
\end{table}

\subsection{Visualizations}

\begin{figure*}
\centering
\includegraphics[width=\linewidth]{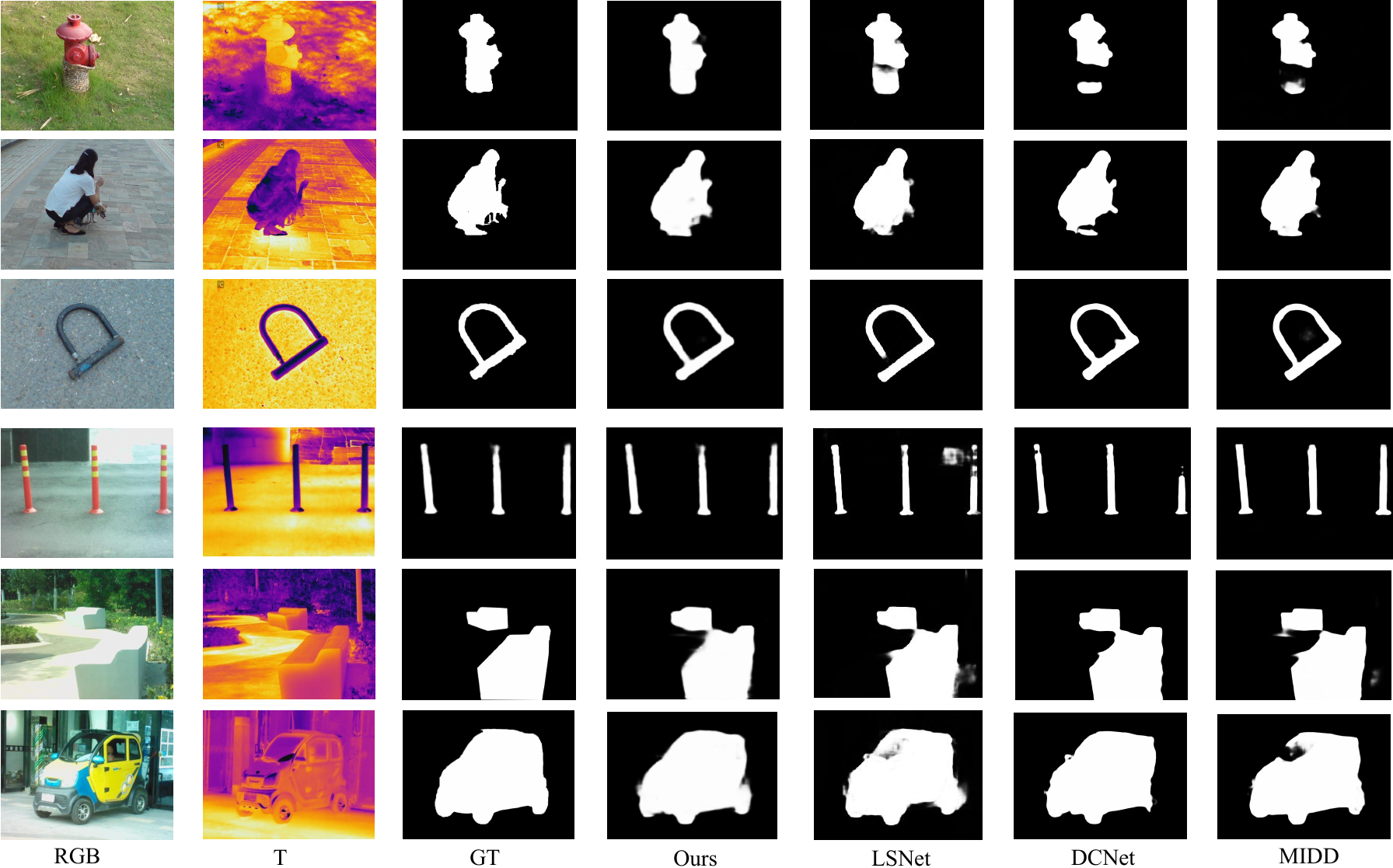}

\caption{Qualitative comparisons with other state-of-the-art methods.}
\label{fig:supp comparisons}
\end{figure*}

Figure \ref{fig:supp comparisons} shows qualitative results between our method and other methods. It should be noted that all the other methods are deep-learning-based and achieve state-of-the-art performance. In some cases such as in row 1, the overlapping area between the salient object and background causes false negatives in the predicted maps generated by other methods. However, our method generates a complete salient object. In order to provide fair visualizations and avoid cherry-picking we also provided more cases where other methods perform well such as the remaining rows. However, our method still outperforms or achieves competitive performance in these qualitative visualizations, especially in some details. In other words, visual maps in Figure \ref{fig:supp comparisons} demonstrate that the proposed method has a better generalization ability that can achieve satisfactory detection results in different scenarios.

\begin{figure*}
\centering
\includegraphics[width=\linewidth]{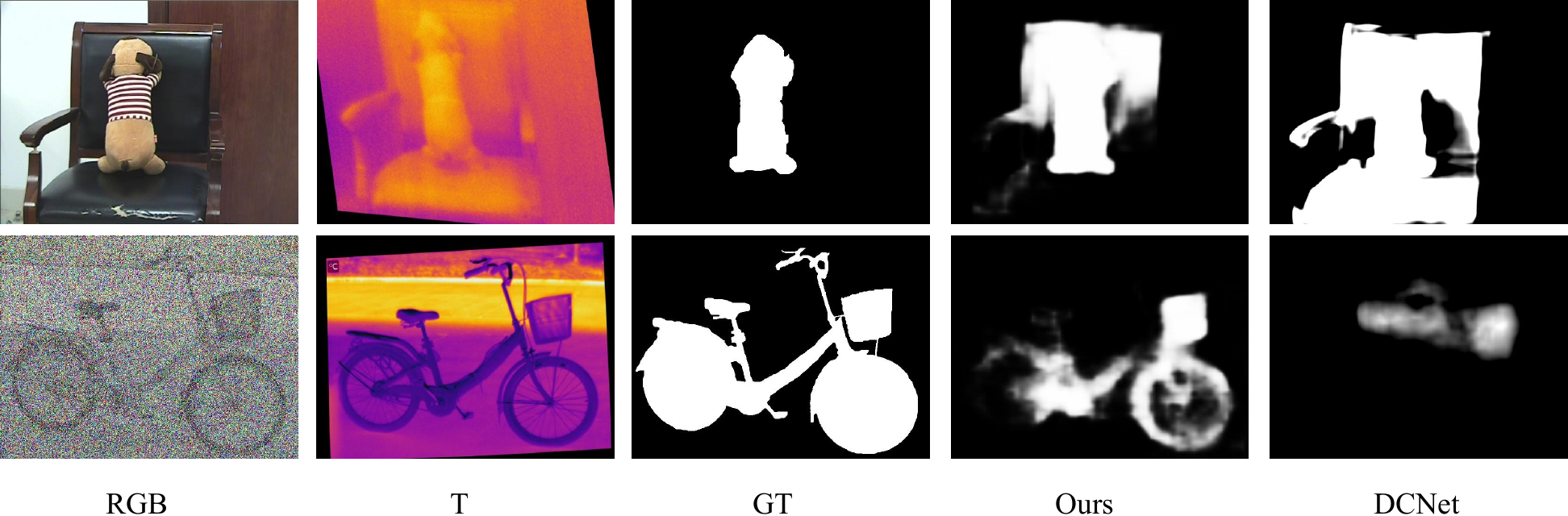}

\caption{Qualitative comparisons on unaligned datasets.}
\label{fig:unaligned supp}
\end{figure*}

In addition to testing with the aligned images, we further provide results for hard unaligned samples and compare the performance with the DCNet \cite{tu2022weakly}, which is designed to tackle the unaligned issue for RGB-T saliency detection. More specifically, row 1 in Figure \ref{fig:unaligned supp} shows the case in which the thermal image has been rotated and stretched. It can be observed that our saliency map has fewer false positives. Row 2 shows the case where not only the images are unaligned but the RGB image is distorted. In this case, our method fairly detects the salient object whereas the DCNet fails. 

\newpage
{\small
\bibliographystyle{ieee}
\bibliography{rgbt_saliency}
}

\end{document}